\documentclass{article}
\pdfoutput=1

\PassOptionsToPackage{numbers,sort&compress}{natbib}
\usepackage[dblblindworkshop,final]{neurips_2026}
\workshoptitle{Foundation Models for Temporal Systems: From Forecasting to World Modeling}

\usepackage[utf8]{inputenc}
\usepackage[T1]{fontenc}
\usepackage{xcolor}
\definecolor{vlink}{HTML}{0F4D92}  %
\usepackage[colorlinks=true,linkcolor=vlink,citecolor=vlink,urlcolor=vlink]{hyperref}
\usepackage{url}
\hypersetup{
  pdftitle={What a Forecasting Leaderboard Can Establish: Resolution Reporting for Zero-Shot Time-Series Models},
  pdfauthor={Arya Sikder, Sriram Pillutla},
  pdfsubject={NeurIPS 2026 Workshop on Foundation Models for Temporal Systems},
  pdfkeywords={time series foundation models, zero-shot forecasting, forecasting leaderboards, benchmark evaluation, evaluation methodology, statistical power, minimum detectable effect, ranking stability, cluster-robust inference}}
\usepackage{graphicx}
\usepackage{booktabs}
\usepackage[labelfont=bf]{caption}
\usepackage{amsmath,amssymb,amsthm}
\usepackage{microtype}
\newtheorem{proposition}{Proposition}

\title{What a Forecasting Leaderboard Can Establish:\\ Resolution Reporting for Zero-Shot Time-Series Models}

\author{%
  Arya Sikder \\
  \texttt{aryasikder17@gmail.com} \\
  \And
  Sriram Pillutla \\
  \texttt{sriramp.iitm@gmail.com}
}

\begin{document}
\maketitle

\begin{abstract}
A foundation model for time series is pretrained once and then asked to forecast series it
has never seen, so the evidence for it is its score on held-out datasets. Such datasets are
scarce, and a few public leaderboards carry most comparisons, yet none of them states how
small a difference its datasets can detect. We call that quantity the \emph{resolution} of
a board and introduce \textbf{Vernier}, a tool that computes it from the files a board
already publishes. Vernier also reports which entries a board cannot separate from its
leader and whether its order changes when unrelated parties submit. GIFT-Eval resolves 6.8
rank places, and its displayed neighbours sit 0.91 apart. The ten places of its top ten hold
fewer than two resolvable steps. \mbox{fev-bench}, Chronos Benchmark~II and TIME show the same
pattern: their resolution is 3.6 to 19 times the gap between neighbours. With each of its 2{,}554 series as an independent source, BOOM is the one board that
resolves its top ten. Below the resolution the order
depends on who else submitted: across four boards, rules that rank against the pool reversed 43 top-ten pairs with
neither entry rerun, in one update in seven. Boards also do not record how much history a model
receives. On 70 configurations, shortening it raised the error of three of four open models,
two of them by 15\%. We release Vernier, which runs this report with one command on any leaderboard that
publishes per-dataset scores: \url{https://github.com/therealram18/vernier}.
\end{abstract}

\begin{figure}[h!]
\centering
\includegraphics[width=\textwidth,alt={Two panels. Left: dot plot of the ten top entries of the GIFT-Eval leaderboard by mean rank behind the leader, spanning 11 rank places, with one bar for the resolution of the board, 6.8 rank places, and one bar for the median gap between neighbours, 0.91. Right: resolution divided by median gap for six forecasting leaderboards on a log axis. GIFT-Eval 7.5, fev-bench 3.6, fev-bench-mini 7.0, Chronos Benchmark II 19, TIME 11 and BOOM 0.63. Only BOOM lies below 1.}]{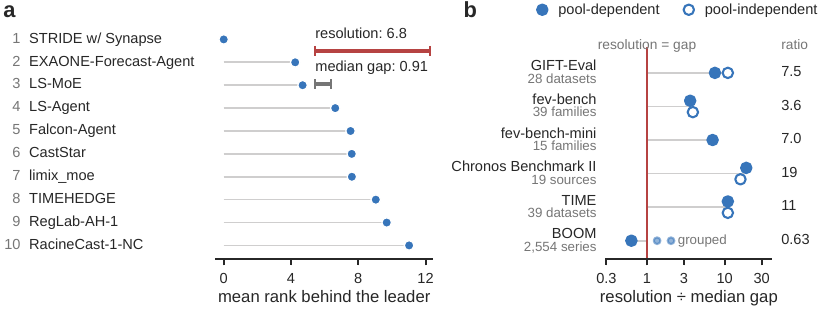}
\caption{\textbf{What a board can resolve.} (a)~GIFT-Eval's top ten by mean rank behind the
leader. The red bar is the board's resolution, the smallest difference that its 28 datasets
can detect. The grey bar is the median gap between neighbours. (b)~Resolution divided by
median gap on six boards, for rules that rank an entry against the pool (filled) and rules
that do not (hollow). Numbers at the right are for each board's displayed rule. Right of the
red line a board displays gaps that it cannot resolve. Only BOOM lies left of it, with each
series as a source (small markers: 36 and 23 groups).}
\label{fig:boards}
\end{figure}

\section{Introduction}

Foundation models for time series are pretrained on large heterogeneous corpora and then
applied to series that were absent from training. The claim that distinguishes them is
transfer to datasets nobody chose in advance, so the evidence has to span many domains,
sampling frequencies and horizons. Independent public forecasting datasets are scarce, so
comparisons run through a few shared boards. The one with the most entries is GIFT-Eval
\citep{gifteval}. It evaluates every submission on the same 97 configurations, each one dataset at one
sampling frequency and forecast horizon. It displays
a strict order over 128 entries, and its per-configuration scores are public. Papers cite a
position on it as evidence of state-of-the-art accuracy \citep{timerouter}.

Every measurement has a resolution, the smallest difference it can separate from noise. A
board measures accuracy with a sample of datasets, and no forecasting board states the
resolution of that sample. GIFT-Eval's 97 configurations come from 28 datasets, four of
which contribute eight configurations each, and the configurations of one dataset rise and
fall together. Its evidence is 28 independent draws, and its top ten spans fewer than two of the steps
they can resolve (Figure~\ref{fig:boards}a).

\paragraph{Vernier.} We treat a board as a measuring instrument and report what it can
resolve. The report answers three questions from a board's published scores alone
(Section~\ref{sec:vernier}). How small a difference can it detect?
Which displayed entries does it separate? Does its order depend on who else submitted? 

\paragraph{Findings.} (i)~GIFT-Eval resolves 6.8 rank places against a median gap of 0.91
between displayed neighbours. Closing that gap would take about 48 times as many
datasets (Section~\ref{sec:boards}). (ii)~The test behind that number holds its level with
28 datasets. It declares a false difference 5.2\% of the time, against 6 to 10\% for a
percentile bootstrap (Appendix~\ref{app:calibration}). (iii)~The
resolution exceeds the median gap on every board built from 15 to 39 public data sources
(Appendix~\ref{app:boards}). It does not on BOOM, whose 2{,}554 series each count as a source. (iv)~Rules
that rank against the pool reversed 43 untouched top-ten pairs on four boards. (v)~No board
records how much history a model receives. Cutting it to 512 steps raised the error of three
of four open models (Section~\ref{sec:protocol}).

\section{Related work}

\paragraph{Uncertainty on leaderboards.} Two forecasting boards already attach uncertainty
to their entries. fev-bench reports paired bootstrap intervals over its tasks
\citep{fevbench}, and TS-Arena collects forecasts before the outcome exists and reports
intervals on its ratings \citep{tsarena}. Outside forecasting, rank intervals exist for
language-model and tabular leaderboards \citep{rankint1, rankint2}, and a quantization audit
fixes the detectable difference before any run \citep{pairedmde}. A mean-rank comparison
between two entries is known to depend on the rest of the pool \citep{benavoli2016,
zhanghardt2024}, and that dependence has been measured on language-model arenas
\citep{arenaperturb}. An interval describes one comparison. No forecasting board states the
smallest difference that its datasets can detect, or tests whether its order depends on the
pool.

\paragraph{Evaluation practice for pretrained forecasters.} Recent work revisits long-horizon evaluation
\citep{positionpaper}, audits pretraining corpora for contamination \citep{tsfmaudit} and
sweeps the history length given to a model \citep{tsfmbench}. We ask what a board must
record for two evaluations of one model to agree.

\section{What Vernier computes}
\label{sec:vernier}

Vernier reads one file, a score for every entry on every evaluation unit (a configuration on
GIFT-Eval, a task on fev-bench), and answers three questions. Appendix~\ref{app:validity} states when each claim holds.

\paragraph{How small a difference can the board detect?} The units are a sample of the
domains a pretrained model might meet, so the difference between two entries carries
sampling error. Units from one source are not independent, so the source is the unit of sampling. We estimate the
standard error (SE) of the mean difference with CR2, a cluster-robust estimator corrected
for few clusters, with Satterthwaite degrees of freedom $\nu$ \citep{bellmccaffrey2002,
imbenskolesar2016}. The \textbf{resolution} of a pair is the smallest true difference that a
5\% test detects with 80\% power, $(t_{0.975,\nu}+t_{0.80,\nu})\,\mathrm{SE}$, the minimum
detectable effect \citep{bloom1995}. A board's resolution is the median over the neighbouring pairs of its top ten.

\paragraph{Which entries cannot be separated from the leader?} A pair \emph{separates} when
the CR2 $t$-test rejects a zero difference at the 5\% level. We report counts uncorrected and after a Holm correction
for multiple comparisons \citep{holm1979}. We repeat every test with a wild cluster
bootstrap, which gives each source's differences one random sign \citep{cameron2008}. We
also run its exact form, which lists every sign pattern. The exact form holds its level for
any number of sources when the two entries are interchangeable within each source
(Appendix~\ref{app:exact}). Entries that do not separate from the leader form
its \textbf{tier}.

\paragraph{Does the order depend on who else submitted?} A mean-rank rule scores an entry
against the rest of the pool (it is \emph{pool-dependent}), so an unrelated submission can
move it. We replay the board's version history and skip any update in which
the evaluation set changed. At each update we take the entries that sat in the displayed top
ten beforehand and whose own result and metadata files are unchanged. We count the pairs
among them that reverse. This is the \textbf{bystander test}.

\section{Results}
\label{sec:boards}

\begin{table}[t]
\caption{\textbf{Vernier on six boards}, each under the rule it sorts by (mean rank for Chronos
Benchmark~II). \emph{Ratio}: resolution over the median gap between displayed neighbours.
\emph{Separated}: neighbouring top-ten pairs that CR2 separates, uncorrected.
\emph{Reversed pairs}: top-ten pairs that reversed with neither entry changed.}
\label{tab:report}
\centering
\footnotesize
\setlength{\tabcolsep}{4.6pt}
\begin{tabular}{@{}lrrlrcrrr@{}}
\toprule
 & & & & & & \multicolumn{2}{c}{Reversed pairs} & \\
\cmidrule(lr){7-8}
Board & Entries & Units & Sources & Ratio & Separated & pool-dep. & pool-indep. & Updates \\
\midrule
GIFT-Eval & 128 & 97 & 28 datasets & 7.5 & 1 of 9 & \textbf{20}$^{a}$ & 0 & 141 \\
fev-bench & 40 & 100 & 39 families & 3.6 & 0 of 9 & \textbf{2} & 0 & 30 \\
fev-bench-mini & 40 & 20 & 15 families & 7.0 & 1 of 9 & -- & -- & -- \\
Chronos Benchmark II & 21 & 27 & 19 sources & 19 & 0 of 9 & \textbf{8} & 0 & 11 \\
TIME & 43 & 98 & 39 datasets & 11 & 0 of 9 & \textbf{13}$^{b}$ & 0 & 44 \\
\midrule
BOOM & 23 & 6{,}750 & 2{,}554 series & \textbf{0.63} & \textbf{9 of 9} & -- & -- & -- \\
\quad grouped by domain & & & 23 groups & 2.04 & 5 of 9 & & & \\
\bottomrule
\end{tabular}

\vspace{2pt}
\parbox{\textwidth}{\scriptsize\centering $^{a}$\,The board's exact tie-break; 18 with tied
ranks averaged. \ $^{b}$\,TIME's MASE rank column; TIME sorts by a geometric mean. \ A dash:
history not replayed. \ Snapshots: GIFT-Eval 2026-09-07, TIME 2026-10-31, the others
2026-11-11.}
\end{table}

\paragraph{Ten displayed places, fewer than two resolvable steps.} GIFT-Eval's 28 datasets
resolve 6.8 rank places. The median gap between displayed neighbours is 0.91, and the top
ten spans 11.0 (Figure~\ref{fig:boards}a). Of the nine neighbouring pairs, the CR2 test
separates one before a Holm correction (places 9 and 10, $p=0.036$). After the correction no
neighbouring pair separates, under CR2, the wild cluster bootstrap or the exact tests. Each
of the nine gaps is smaller than its own pair's resolution. Closing the median gap would take
about 48 times as many datasets (Appendix~\ref{app:gaps}).

\paragraph{The same report on five more boards.} We ran Vernier on five more forecasting
boards that publish per-entry, per-unit scores: fev-bench and its 20-task subset
\citep{fevbench}, Chronos Benchmark~II \citep{chronos}, TIME \citep{time} and BOOM
\citep{boom} (Table~\ref{tab:report}). On every board built from 15 to 39 public data sources the
resolution is 3.6 to 19 times the median gap. At most one of nine neighbouring pairs
separates under mean rank, win rate, skill score or geometric mean. With every family that shares a
data source with GIFT-Eval removed, fev-bench still separates no neighbouring pair. On ten
subsets of GIFT-Eval that its app shows as tabs, the ratio is 6.9 to 43 and no neighbouring
pair separates. In November 2026 the newly released Timestral \citep{timestral} sits 11th, 3rd and 4th
on GIFT-Eval, TIME and fev-bench, and no board separates it from either neighbour. BOOM is
the board that resolves. With each of its 2{,}554 series as a source, its resolution is 0.63 of
its median gap, and all nine neighbours separate. Grouped into its 23 domains, five of the
nine separate: independent sources set the resolution, not units.

\paragraph{Below the resolution, the pool sets the order.} A mean-rank rule ranks an entry
against every other entry on each configuration, so a new arrival displaces entries that
nobody reran. The displacement is small, and it reverses a pair whose gap is smaller. In the week to 2026-09-07 GIFT-Eval took three submissions from
three unrelated organisations and changed nothing else. The order of Falcon-Agent,
CastStar and limix\_moe reversed completely (Appendix~\ref{app:replay}). On fev-bench,
second and third place (ChakraTS and TimesFM-3) reversed on 2026-10-14 when an unrelated
entry arrived. On the four boards whose history we could replay, pool-dependent rules
reversed 43 untouched top-ten pairs, in 31 of 226 updates (Tables~\ref{tab:report} and~\ref{tab:allboards}).
GIFT-Eval reversed 8 more in the ten weeks after the snapshot. The cause is the rule. The
geometric mean of scaled error, an entry's error divided by that of a seasonal-naive
baseline, uses only the entry's own scores (it is \emph{pool-independent}), so no arrival
can reverse an untouched pair. Below the resolution, a displayed order also records who else submitted.

\paragraph{Recommendations based on tiers.} All ten entries at the top of GIFT-Eval are
agentic pipelines, systems that select or combine several forecasters per series. Nine of
the ten declare no reproduction code. The strict order, taken at face value, recommends a
system a practitioner cannot rebuild. Under the displayed rule the leading tier has three
entries. Under the pool-independent rule it has 17 entries under CR2, and 16 under a signed-rank test
that holds its level on that rule. TimesFM-3, a pretrained model released with weights, is
in both (Appendix~\ref{app:tiers}). The recommendation
changes with the rule, so the evidence does not settle the choice between an agentic
pipeline and a pretrained model.

\section{Varying the history}
\label{sec:protocol}

Resampling published scores cannot expose sensitivity to how an evaluation was set up,
because score files record no settings. The choice left most open is the length of history
a model receives. We ran four open models at two history lengths, inference only: Chronos-Bolt-base
\citep{chronos}, Moirai-2.0-R-small \citep{moirai}, TimesFM-2.5-200M \citep{timesfm} and
Toto-Open-Base-1.0 \citep{toto}. The runs cover all 70 GIFT-Eval configurations whose median
series holds at least 1{,}024 steps, from 18 datasets in 14 sources. The
first length is the value in GIFT-Eval's own reference notebook for that model, and the
second is 512 steps. Twelve configurations, chosen by a rule fixed in advance, form a discovery set, and the
other 58 a confirmation set. The estimate and the tests were fixed in advance
(Appendix~\ref{app:history}).

\begin{table}[t]
\caption{\textbf{Error at 512 steps of history against the documented length.} Change in
the geometric mean of scaled error (positive is worse), with its 95\% interval and
Holm-corrected $p$-value. Bold separates from zero. The last two columns split the 70
configurations. $\dagger$:~declares test-data leakage.}
\label{tab:protocol}
\centering
\footnotesize
\setlength{\tabcolsep}{6pt}
\begin{tabular}{@{}lcccc@{}}
\toprule
Model & 70 configurations [95\% interval] & $p$ after Holm & Confirmation, 58 & Discovery, 12 \\
\midrule
TimesFM-2.5-200M & \textbf{+15.3\%} [+4.1, +32.6] & \textbf{0.001} & +14.0\% & +22.2\% \\
Toto-Open-Base-1.0 & \textbf{+15.3\%} [+4.0, +27.1] & \textbf{0.001} & +13.9\% & +22.0\% \\
Moirai-2.0-R-small & \textbf{+6.2\%} [+1.6, +10.4] & \textbf{0.010} & +6.1\% & +6.8\% \\
Chronos-Bolt-base\,$^\dagger$ & +2.7\% [$-$0.9, +6.2] & 0.11 & +3.7\% & $-$2.0\% \\
\midrule
Spread of the displayed top ten & 11.8\% & & 12.7\% & 9.3\% \\
\bottomrule
\end{tabular}
\end{table}

\paragraph{History length moves three of four models.} On the 70 configurations the shorter
history raised the error of TimesFM-2.5 and of Toto by 15.3\% and that of Moirai-2.0 by
6.2\%. All three separate from zero after a Holm correction (Table~\ref{tab:protocol}), and
they separate again on the confirmation set. For TimesFM-2.5 and Toto the error rises on 64
of the 70 configurations and in 13 of the 14 sources (one-sided sign test over sources, $p<0.001$). No
effect is established for Chronos-Bolt. The estimates for TimesFM-2.5 and Toto lie above the
11.8\% spread of the displayed top ten. Their intervals start
near 4\%, so a test against that spread does not reject. On the discovery set the
newer TimesFM-3 and Timestral-209M rise by 31.8\% and 28.9\%. A documented length is also
an upper limit: at the median evaluation window, 13 of the 48 discovery-set runs received
less history than documented (Figure~\ref{fig:received}). Sampling is a second unrecorded
setting. With one point forecast per series, Toto misses the board's rows by 5.5\%. With
the reference notebook's setting, the median of 256 samples, it reproduces them to 0.02\%.
We attribute the gap mainly to the sample count.
A board should record the history received and the sampling setting.

\section{Conclusion}

Zero-shot forecasting boards display an order that is finer than their datasets can support.
Below the resolution, a place reflects who else submitted and how the model was run as much
as how well it forecasts. Shared boards remain the right way to compare these models, read
at the precision they have. A board that reports its resolution, ranks with a
pool-independent rule, shows tiers and records how each model was run loses nothing that it
could measure. What no report can add is evidence. With each of its 2{,}554 series as an independent source, BOOM resolves its top ten, and
the other boards cannot with a few dozen datasets. Better
comparisons will come from more independent data, not from more submissions to the same
datasets.

\bibliographystyle{unsrtnat}
\bibliography{references}
\clearpage
\appendix
\raggedbottom
\setlength{\textfloatsep}{12pt plus 2pt minus 2pt}
\setlength{\floatsep}{12pt plus 2pt minus 2pt}
\setlength{\intextsep}{12pt plus 2pt minus 2pt}

\begin{center}
{\LARGE\bfseries Appendix}
\end{center}
\vspace{4pt}

\noindent Appendix~\ref{app:method} gives the method and shows that its test holds its level
with 28 datasets. Appendix~\ref{app:gifteval} gives the full results on GIFT-Eval,
Appendix~\ref{app:boards} the five other boards, and Appendix~\ref{app:history} the
history-length experiment and the disclosure checklist.

\section{Method and validity}
\label{app:method}

\subsection{Details of the computation}

Section~\ref{sec:vernier} defines the report. For a displayed statistic and two entries,
the difference on one evaluation unit is a difference of ranks under a mean-rank rule and of
log scaled errors under a geometric mean. The displayed gap is the mean of these
differences. The scaled error of an entry on a GIFT-Eval configuration is its MASE divided
by the Seasonal Naive baseline's MASE on the same configuration.

\paragraph{CR2 and its degrees of freedom.} Sources are the clusters. The working model
treats the units of one source as equally correlated, with the pair's own estimated
correlation floored at zero. The degrees of freedom $\nu$ come from Satterthwaite's
approximation under that model \citep{bellmccaffrey2002, imbenskolesar2016}. Over the 45
top-ten pairs of GIFT-Eval $\nu$ runs from 11.41 to 17.07 under mean rank on MASE. The
multiplier $t_{0.975,\nu}+t_{0.80,\nu}$ of the resolution is then about 3.0, against 2.80 in
the large-sample limit. Holm corrections run over the nine neighbouring pairs when we count
neighbours and over the 45 pairs of the top ten when we count all pairs. The leading tier of
a rule holds the leader and every entry that the uncorrected test does not separate from it,
among the first 40 entries in that rule's own order.

\paragraph{Agreement with reference software.} On the 45 GIFT-Eval pairs under mean rank on
MASE, the R package clubSandwich 0.7.0 \citep{pustejovsky2018} gives the same CR2 standard
errors and, under independent errors, the same degrees of freedom. The largest relative
difference is $2.5\times10^{-15}$. The R package dfadjust 1.1.0 computes the degrees of freedom of
\citet{imbenskolesar2016}. Given its correlation estimate our formula reproduces its values
to $1.5\times10^{-14}$, and with its correlation floored at zero it gives exactly our
counts. With its default, which does not floor the correlation, the same 1 of 9 neighbours
and 12 of 45 pairs separate, and no neighbour after the Holm correction. The Python package wildboottest 0.3.2 gives the same $t$ statistic to
$9\times10^{-16}$, and all 45 of its $p$-values lie within 0.0048 of ours.

\subsection{Does the test hold its level with 28 datasets?}
\label{app:calibration}

A test over 28 clusters can reject too often. We measured how often four procedures declare
a difference when there is none (Table~\ref{tab:calibration}). The first two are CR2 and the
wild cluster bootstrap \citep{cameron2008, canay2021}. The third is a $t$-test on the 28
dataset means, which weights every dataset equally. The fourth is a percentile bootstrap,
under which a pair separates when its 95\% intervals from resampling configurations and
from resampling whole datasets both exclude zero. CR2 was named in advance as the primary procedure, to be replaced only if it
fell outside 4 to 6\% on the resampled real pairs under mean rank on MASE. Its rate there is
5.2\%.

\begin{table}[!htb]
\caption{\textbf{False alarms at a nominal 5\% level}, in percent. The first seven rows
simulate 4{,}000 boards with GIFT-Eval's layout, 28 datasets with 1 to 8 configurations each
(standard error about 0.35 points). The last two rows resample the 28 datasets 1{,}000 times
for each of the 45 top-ten pairs.}
\label{tab:calibration}
\centering
\footnotesize
\begin{tabular}{@{}lcccc@{}}
\toprule
Data with no true difference & CR2 & Wild bootstrap & $t$ on means & Percentile \\
\midrule
Configurations independent & 5.0 & 5.0 & 4.9 & 4.4 \\
Mild within-dataset correlation (0.1) & 5.0 & 4.6 & 4.9 & 6.0 \\
Moderate correlation (0.3) & 4.8 & 4.7 & 5.5 & 7.0 \\
Moderate correlation, heavy tails & 4.4 & 4.8 & 4.7 & 8.1 \\
Moderate, effect grows with dataset size & 5.8 & 5.4 & 4.9 & 8.5 \\
Moderate, strongly lopsided dataset effects & 5.8 & 5.9 & 5.3 & 8.4 \\
Strong correlation (0.5), strongly lopsided & 6.8 & 7.0 & 5.4 & 9.5 \\
\midrule
Real pairs resampled, mean rank & \textbf{5.2} & 5.4 & 6.0 & 6.1 \\
Real pairs resampled, geometric mean & 15.7 & 15.5 & 8.4 & 16.2 \\
\bottomrule
\end{tabular}
\end{table}

The real mean-rank pairs have a within-dataset correlation of 0 to 0.51 with a median of
0.14, so the mild and moderate rows of Table~\ref{tab:calibration} describe this board. CR2
holds 4.4 to 5.8\% in every setting but the most extreme one, where it reaches 6.8\%. The
percentile bootstrap rejects 6.0 to 9.5\% whenever configurations within a dataset are
correlated, and its independent-configuration scheme alone rejects 10 to 35\%.

\paragraph{The geometric mean.} On resampled real geometric-mean pairs these four procedures
reject too often (Table~\ref{tab:calibration}, last row). The rate is about 16\% for CR2,
the wild bootstrap and the percentile bootstrap, and 8.4\% for the $t$-test on dataset
means. The cause
is concentration. In 28 of the 45 pairs one dataset supplies more than half of the variance
estimate, with a median share of 61\% and a maximum of 93\%. That dataset is most often
\texttt{solar}, which contributes 8 configurations. Under mean rank the median share is 31\%
and 3 pairs exceed one half, because ranks are bounded. On TIME, one dataset supplies more
than half of the variance in 18 of 45 pairs under the geometric mean and in 7 under mean
rank. A pool-independent rule removes the dependence on other entries and concentrates each
comparison on few sources. A board that adopts one should publish, beside each comparison,
the largest share of the variance that one source supplies.

\paragraph{A test that holds its level on the geometric mean.} Four candidate tests, their order and a pass rule were fixed in advance. A test passes
when, over 1{,}000 resamples of the 28 datasets per pair, its pooled false-alarm rate lies in
4 to 6\%, at most 2 of the 45 pairs exceed 7.5\% and none exceeds 10\%. One candidate passes, the exact Wilcoxon signed-rank
test with datasets as the units. It averages a pair's differences in log scaled error within
each dataset, ranks the 28 averages by absolute size, and compares the signed sum of ranks
with all $2^{28}$ sign patterns. It is the test of Proposition~\ref{prop:exact} with a
rank statistic. Its pooled rate is 5.6\% and its worst pair 7.5\%. The worst pair of CR2 in the same
resamples is 37.0\%. Sign-flip tests on the dataset means, with
and without clipping, fail at 8.5 to 17.5\%. The signed-rank test asks whether the typical
dataset favours one entry, with each dataset counted once. It does not test the board's
configuration-weighted mean, and no test of that mean held its level at 28 datasets. On the
displayed top ten it separates 1 of 9 neighbouring pairs and 12 of 45 pairs under the
geometric mean. After a Holm correction it separates no neighbour and 1 of the 45. The conclusion under the
pool-independent rule is therefore the same: no neighbouring pair separates after
correction.

\subsection{An exact test for interchangeable entries}
\label{app:exact}

Let $D_g$ hold the differences between two entries on the units of source $g$, and let $T$
be any statistic of $D=(D_1,\dots,D_G)$. Swapping the labels of the two entries inside
source $g$ replaces $D_g$ by $-D_g$. Under a rank rule with tied ranks averaged, the ranks
of all other entries stay as they were.

\begin{proposition}
\label{prop:exact}
Suppose the two entries are interchangeable within every source: the joint distribution of
$(D_1,\dots,D_G)$ is unchanged when any subset of the $D_g$ change sign. Let
$p(D)=2^{-G}\,\#\{s\in\{-1,+1\}^G : T(s_1D_1,\dots,s_GD_G)\ge T(D)\}$. Then
$\Pr\{p(D)\le\alpha\}\le\alpha$ for every $\alpha$, every number of sources $G\ge 1$ and
every set of source sizes.
\end{proposition}

\begin{proof}
The $2^G$ sign changes form a group, and the null distribution is invariant under it. The
bound is the standard property of a randomization test \citep[Chapter~17]{lehmannromano2022}.
\end{proof}

With 28 datasets all $2^{28}$ sign patterns can be listed, so the test needs no random
draws. We ran it with two statistics, the mean difference over its clustered standard error
and the plain sum of the differences. The wild cluster bootstrap with the null imposed is
the Monte Carlo form of the first. Each draw gives every dataset one sign with equal
probability. The $p$-value $(1+b)/(1+B)$ counts the observed statistic as one draw, so
it obeys the same bound for any number of draws $B$. We used $B=99{,}999$. On every pair
its $p$-value is within 0.0035 of the exact one, and no neighbour or Holm decision differs
(Table~\ref{tab:counts}). The guarantee needs no large number of sources, no equal sizes and
no thin tails. Table~\ref{tab:calibration} agrees with it: the wild bootstrap rejects 4.6 to
5.4\% in the five settings without lopsided dataset effects, and 5.9 and 7.0\% in the two
settings with them.

\subsection{When each claim holds}
\label{app:validity}

\paragraph{What the tests answer.} They answer whether the displayed order would survive an
equally reasonable draw of datasets from the population this board samples. On the fixed
evaluation set the published means are exact.

\paragraph{The unit of sampling.} Datasets are the independent unit, and configurations of
one dataset are correlated. Rolling windows and repeat runs vary inside a configuration's
score, so the standard error between datasets already contains their variation. A setting
such as the history length is the same on every dataset, so no standard error contains it
(Section~\ref{sec:protocol}).

\paragraph{How many more datasets.} The figure of 48 times as many datasets holds when new
datasets vary as the current ones do. On the 25 fev-bench families that share no data source
with GIFT-Eval the noise scale is 1.25 times as large, so the figure is a lower estimate
(Appendix~\ref{app:boards}).

\paragraph{The history experiment.} The sign-flip test on source totals is exact when,
with no true effect, the distribution of each source total is symmetric about zero. The changes
are skewed toward a rise, which makes the test conservative for a rise. A false claim of a
rise occurs in at most 1\% of draws for the three models with an effect
(Appendix~\ref{app:history}).
Each difference compares two runs of ours on one configuration, so it does not depend on any
board row.

\section{GIFT-Eval in full}
\label{app:gifteval}

\subsection{Neighbours, pairs and resolution}
\label{app:gaps}

\begin{table}[!htb]
\caption{\textbf{The nine neighbouring pairs of GIFT-Eval's top ten} under mean rank on
MASE, at the snapshot of 2026-09-07. Figure~\ref{fig:boards}a names the entries. The exact
test lists all $2^{28}$ sign patterns, and Holm corrects over the nine pairs. The board's
resolution is the median of the last column, 6.80.}
\label{tab:adjacent}
\centering
\footnotesize
\setlength{\tabcolsep}{5pt}
\begin{tabular}{@{}lrrrcrrrr@{}}
\toprule
Places & Gap & CR2 SE & $\nu$ & 95\% interval & $p$, CR2 & $p$, exact & Holm $p$ & Resolution \\
\midrule
1--2  & 4.25 & 3.05 & 11.9 & [$-$2.40, 10.89] & 0.1889 & 0.1760 & 1 & 9.31 \\
2--3  & 0.44 & 3.30 & 13.0 & [$-$6.68, \phantom{1}7.57] & 0.8952 & 0.8940 & 1 & 10.00 \\
3--4  & 1.94 & 0.99 & 17.1 & [$-$0.15, \phantom{1}4.02] & 0.0665 & 0.0624 & 0.53 & 2.94 \\
4--5  & 0.91 & 2.24 & 17.1 & [$-$3.82, \phantom{1}5.64] & 0.6909 & 0.6906 & 1 & 6.67 \\
5--6  & 0.07 & 2.29 & 17.1 & [$-$4.75, \phantom{1}4.90] & 0.9752 & 0.9774 & 1 & 6.80 \\
6--7  & 0.01 & 1.42 & 17.1 & [$-$2.98, \phantom{1}3.00] & 0.9943 & 0.9998 & 1 & 4.21 \\
7--8  & 1.42 & 2.62 & 14.4 & [$-$4.17, \phantom{1}7.02] & 0.5947 & 0.5883 & 1 & 7.86 \\
8--9  & 0.65 & 3.35 & 12.4 & [$-$6.63, \phantom{1}7.93] & 0.8496 & 0.8523 & 1 & 10.20 \\
9--10 & 1.33 & 0.59 & 17.1 & [\phantom{$-$}0.09, \phantom{1}2.57] & \textbf{0.0364} & \textbf{0.0404} & 0.33 & 1.74 \\
\bottomrule
\end{tabular}
\end{table}

The nine gaps of Table~\ref{tab:adjacent} sum to 11.02 rank places, the span of the top ten,
and their median is 0.91. Each gap is smaller than its own pair's resolution. Only the pair
at places 9 and 10, RegLab-AH-1 and RacineCast-1-NC, separates before correction. Its $p$-value is 0.036
under CR2, 0.041 under the wild cluster bootstrap, 0.040 under the exact test with the
bootstrap's statistic and 0.048 with the plain sum. After a Holm correction over the nine
neighbours its $p$-value is 0.33.

\begin{table}[!htb]
\caption{\textbf{Separated pairs of GIFT-Eval's top ten}, by rule and test. Each cell gives
the count before / after a Holm correction, among the 9 neighbouring pairs and among all 45.
These tests do not hold their level on the geometric mean, where the signed-rank test of
Appendix~\ref{app:calibration} replaces them.}
\label{tab:counts}
\centering
\footnotesize
\setlength{\tabcolsep}{4.5pt}
\begin{tabular}{@{}lcccccccc@{}}
\toprule
 & \multicolumn{2}{c}{CR2} & \multicolumn{2}{c}{Wild bootstrap} & \multicolumn{2}{c}{Exact, same statistic} & \multicolumn{2}{c}{Exact, plain sum} \\
\cmidrule(lr){2-3}\cmidrule(lr){4-5}\cmidrule(lr){6-7}\cmidrule(l){8-9}
Rule & of 9 & of 45 & of 9 & of 45 & of 9 & of 45 & of 9 & of 45 \\
\midrule
Mean rank on MASE (displayed) & 1 / 0 & 12 / 0 & 1 / 0 & 13 / 1 & 1 / 0 & 13 / 1 & 1 / 0 & 12 / 1 \\
Mean rank on CRPS (earlier rule) & 0 / 0 & 9 / 0 & 0 / 0 & 12 / 0 & 0 / 0 & 11 / 0 & 0 / 0 & 12 / 0 \\
Geometric mean of scaled error & 0 / 0 & 10 / 0 & 0 / 0 & 19 / 4 & 0 / 0 & 19 / 4 & 0 / 0 & 18 / 4 \\
\bottomrule
\end{tabular}
\end{table}

Under both mean-rank rules no neighbouring pair separates after the Holm correction, under
any of the four tests (Table~\ref{tab:counts}). Under mean rank on MASE the exact test
separates one of the 45 pairs after the correction, LS-MoE against RacineCast-1-NC at places
3 and 10 (adjusted $p=0.017$, and $0.057$ under CR2).

\paragraph{Resolution.} The resolution is 6.80 rank places under mean rank on MASE. It is
7.38 under mean rank on CRPS, against a median gap of 0.67, and 0.0497 on the log scale
under the geometric mean, against 0.0045. The number of datasets needed is computed for each
neighbouring pair as 28 times the squared ratio of 2.80 standard errors to its gap. Its
median over the nine pairs, at places 4 and 5, is 1{,}342 datasets, 48 times the 28. The
resolution is stable. It is 6.75 rank places with the 28 datasets grouped into 19 sources
(the M4, ETT, bizitobs and bitbrains subsets merged), and between 6.17 and 7.43 with any one
dataset left out. The standard error scales as that calculation assumes. Relative to its value at 28
datasets, the CR2 standard error times the square root of the number of datasets has a
median of 0.991 at 14 datasets and 1.003 at 21. In simulation CR2 detects a difference at a
pair's own resolution 79 to 93\% of the time. With the board's 6.8 planted on all nine
neighbouring pairs, it detects 75\% on the resampled real data. The nine pairs have
resolutions from 1.7 to 10.2, and their standard errors predict 75.2\%.

\subsection{Tiers and the top of the board}
\label{app:tiers}

\begin{table}[!htb]
\caption{\textbf{The leading tier of GIFT-Eval} under each rule and test, in that rule's own order.
A tier holds the leader and every entry that the uncorrected test does not separate from it.
TimesFM-3, in bold, is a pretrained model released with weights.}
\label{tab:tiers}
\centering
\footnotesize
\setlength{\tabcolsep}{4.5pt}
\begin{tabular}{@{}llcp{0.535\textwidth}@{}}
\toprule
Rule & Test & Size & Members \\
\midrule
Mean rank on MASE & CR2, wild bootstrap & 3 & STRIDE w/ Synapse, EXAONE-Forecast-Agent, LS-MoE \\
Mean rank on CRPS & CR2 & 3 & STRIDE w/ Synapse, EXAONE-Forecast-Agent, LS-MoE \\
Geometric mean & CR2 & 17 & \raggedright EXAONE-Forecast-Agent, STRIDE w/ Synapse, TIMEHEDGE,
LS-MoE, LS-Agent, CastStar, TurkForecast-FM-Chronos2-LoRA-v1, limix\_moe, ForecastMate,
RegLab-AH-1, FastCastrage, Falcon-Agent, Falcon-2.0, \textbf{TimesFM-3},
Taichu-TimeSeries-Agent, RAES-Conductance-Ensemble-V, STRIDE (+Chronos-2) \tabularnewline
 & Signed-rank & 16 & \raggedright The CR2 members without ForecastMate, FastCastrage,
Taichu-TimeSeries-Agent and STRIDE (+Chronos-2), and with RacineCast-1, RacineCast-1-NC and
Toto-2.0-FnF \tabularnewline
\bottomrule
\end{tabular}
\end{table}

Under mean rank on MASE, CR2 and the wild cluster bootstrap give the same three entries
(Table~\ref{tab:tiers}). Under the geometric mean the leader is EXAONE-Forecast-Agent, and
both tests give 17 entries. The signed-rank test on dataset means, which holds its level on
that rule (Appendix~\ref{app:calibration}), gives 16. TimesFM-3 is in the geometric-mean tier
under all three tests. The two rules also disagree below the top ten. One entry is 43rd
under mean rank on MASE and 7th under the geometric mean, and two more differ by 21 and 15
places.

\paragraph{The top ten.} All ten displayed entries are agentic pipelines. Nine of them
declare no reproduction code, against 55 of all 130 registered entries. Seventeen entries
declare test-data leakage, and none of them is in the top ten. The board displays 128 of the
130, because two are lost when its app joins result files to declarations by name.

\subsection{The bystander test}
\label{app:replay}

\begin{figure}[!htb]
\centering
\includegraphics[width=\linewidth,alt={Two panels. Left: cumulative count of top-ten pairs on the GIFT-Eval leaderboard that reversed with neither entry rerun, from late 2024 to September 2026. Under mean rank on MASE the count reaches 20. Under the geometric mean it stays at 0. Right: places of seven top-ten entries before and after three entries arrived on 2026-09-07. Three of the seven reverse their order.}]{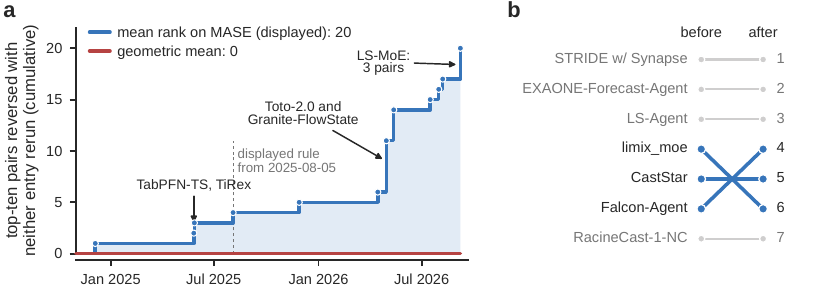}
\caption{\textbf{The bystander test on GIFT-Eval.} (a)~Top-ten pairs that reversed with
neither entry rerun, counted over 141 updates. Mean rank on MASE reverses 20, and the
geometric mean none. Labels name the arrivals behind three steps. (b)~The seven top-ten
entries already on the board, before and after three entries arrived on 2026-09-07.
limix\_moe, CastStar and Falcon-Agent reverse their order.}
\label{fig:timeline}
\end{figure}

Replayed under mean rank on MASE with the board's exact tie-break, 14 of GIFT-Eval's 141
usable updates reversed 20 top-ten pairs with neither entry rerun
(Figure~\ref{fig:timeline}a, Table~\ref{tab:replay}). With tied ranks averaged the count is
18, also in 14 updates. The geometric mean reverses none, by definition: an entry's value
does not change when another entry arrives. One vendor's two model versions, RacineCast-1
and RacineCast-1-NC, reversed on 2026-07-31 and reversed back a week later. The board sorted
by mean rank on CRPS until 2025-08-05. Ten of the 14 reversing updates, with 16 of the 20
pairs, come after that date.

\begin{table}[!htb]
\caption{\textbf{The 20 reversed pairs on GIFT-Eval.} Top-ten pairs that reversed under mean
rank on MASE with neither entry's files changed, in 14 of the 141 usable updates up to
2026-09-07. Each pair is listed in its order before the update.}
\label{tab:replay}
\centering
\scriptsize
\setlength{\tabcolsep}{5pt}
\begin{tabular}{@{}llll@{}}
\toprule
Update ends & Arrived & Pair that reversed & Places before $\rightarrow$ after \\
\midrule
2024-12-05 & chronos\_bolt\_base, chronos\_bolt\_small & Moirai\_small, TFT & 8, 9 $\rightarrow$ 11, 10 \\
2025-05-27 & TabPFN-TS & YingLong\_110m, chronos\_bolt\_small & 7, 8 $\rightarrow$ 8, 7 \\
2025-05-28 & TiRex & YingLong\_300m, timesfm\_2\_0\_500m & 5, 6 $\rightarrow$ 7, 6 \\
2025-08-04 & TimeCopilot & chronos\_bolt\_small, YingLong\_110m & 8, 9 $\rightarrow$ 10, 9 \\
2025-11-28 & Xihe-ultra & Kairos-1.0, Migas-1.0 & 4, 5 $\rightarrow$ 5, 4 \\
2026-04-15 & VISIT-2.0 & Synapse, Migas-1.0 & 7, 8 $\rightarrow$ 8, 7 \\
2026-04-30 & Toto-2.0, five sizes & Migas-1.0, Synapse & 7, 8 $\rightarrow$ 10, 7 \\
 & & Migas-1.0, ZooCast-Top1 & 7, 9 $\rightarrow$ 10, 9 \\
2026-04-30 & Granite-FlowState-r1.1 & Toto-2.0-2.5B, ZooCast-Top1 & 8, 9 $\rightarrow$ 10, 9 \\
 & & Toto-2.0-2.5B, Migas-1.0 & 8, 10 $\rightarrow$ 10, 8 \\
 & & ZooCast-Top1, Migas-1.0 & 9, 10 $\rightarrow$ 9, 8 \\
2026-05-13 & RAES-Conductance-Ensemble & ZooCast-Top1, Toto-2.0-2.5B & 9, 10 $\rightarrow$ 10, 9 \\
2026-05-13 & Toto-2.0-2.5B-FT, Toto-2.0-FnF & TimeCopilot, Migas-1.0 & 7, 9 $\rightarrow$ 11, 9 \\
 & & Synapse, Migas-1.0 & 8, 9 $\rightarrow$ 10, 9 \\
2026-07-16 & Falcon-Agent & Prism, Toto-2.0-FnF & 3, 4 $\rightarrow$ 5, 4 \\
2026-07-31 & limix\_moe & RacineCast-1-NC, RacineCast-1 & 3, 4 $\rightarrow$ 5, 4 \\
2026-08-07 & LS-Agent & RacineCast-1, RacineCast-1-NC & 6, 7 $\rightarrow$ 8, 7 \\
2026-09-07 & LS-MoE & limix\_moe, CastStar & 4, 5 $\rightarrow$ 7, 6 \\
 & & limix\_moe, Falcon-Agent & 4, 6 $\rightarrow$ 7, 5 \\
 & & CastStar, Falcon-Agent & 5, 6 $\rightarrow$ 6, 5 \\
\bottomrule
\end{tabular}
\end{table}

\paragraph{One update.} RegLab-AH-1, TIMEHEDGE and LS-MoE all arrived on 2026-09-07. Without
them limix\_moe, CastStar and Falcon-Agent hold mean ranks of 20.59, 20.76 and 20.85. With
them the three hold 22.24, 22.23 and 22.15, so their order reverses completely
(Figure~\ref{fig:timeline}b). The three pairs reverse at the arrival of LS-MoE, the last row
of Table~\ref{tab:replay}. The effect is general. Of the 131 entries with result files, 121
lie outside the top ten. Removing any one of them reorders the top ten in 28 cases. Adding a near-copy of one displayed
entry, a duplicate with every error multiplied by 1.000001, reorders it in 42 of 128 cases.

\paragraph{After the snapshot.} Between 2026-09-07 and 2026-11-13 GIFT-Eval took 38 more
updates, and they reversed 8 more untouched top-ten pairs, in 8 updates. Seven of the eight
are limix\_moe and EXAONE-Forecast-Agent exchanging places as other entries arrived. At the
later date the board displays 170 entries. Its resolution is 17.28 rank places against a
median gap of 0.48, a ratio of 36 against 7.5 at the snapshot, and no neighbouring pair
separates.

\section{The same report on other boards}
\label{app:boards}

\paragraph{Boards and reproduction.} We ran Vernier on every forecasting board whose
per-entry, per-unit scores are public: fev-bench and its 20-task subset
\citep{fevbench}, Chronos Benchmark~II \citep{chronos}, BOOM \citep{boom} and TIME
\citep{time}. The units of each board, and what counts as an independent source on it, were
fixed before any of its scores was analysed. Each analysis first reproduced the board's own
numbers. fev-bench, fev-bench-mini and Chronos Benchmark~II match the board's own
leaderboard function exactly. BOOM matches its published table in all six columns at the
three decimals it publishes. TIME's recomputed order equals the board's displayed order in
all 45 states of its history. GIFT-Eval is pinned on 2026-09-07, TIME on 2026-10-31 and the
other boards on 2026-11-11, and the release holds every pinned snapshot. Chronos Benchmark~I, OpenTS/TFB, ProbTS, Time-Series-Library and
the Monash archive publish no per-entry, per-unit file. TSFM-Bench publishes code without
result files, and TS-Arena sorts by a rating whose update formula is not published, so its
order cannot be reproduced.

\begin{table}[!htb]
\caption{\textbf{Vernier on every board and rule.} \emph{Pool}: the rule ranks an entry
against the other entries. \emph{Ratio}: resolution over median gap. \emph{Separated}:
neighbouring top-ten pairs, uncorrected, under CR2 / wild bootstrap / percentile bootstrap.
\emph{Reversed}: untouched top-ten pairs that reversed. Skill scores and Chronos geometric
means are log error ratios times 100. The three tests do not hold their level on
geometric-mean rows (Appendix~\ref{app:calibration}).}
\label{tab:allboards}
\centering
\footnotesize
\setlength{\tabcolsep}{4.2pt}
\begin{tabular}{@{}llrrrcrl@{}}
\toprule
Board and rule & Pool & Resolution & Median gap & Ratio & Separated & Reversed & Updates \\
\midrule
GIFT-Eval, mean rank on MASE & yes & 6.80 & 0.91 & 7.5 & 1 / 1 / 0 & 20$^{a}$ & 14 of 141 \\
GIFT-Eval, geometric mean (log) & no & 0.0497 & 0.0045 & 11 & 0 / 0 / 2 & 0 & 0 of 141 \\
fev-bench, win rate (points) & yes & 7.51 & 2.08 & 3.6 & 0 / 0 / 0 & 2 & 2 of 30 \\
fev-bench, skill score & no & 3.20 & 0.83 & 3.9 & 0 / 0 / 0 & 0 & 0 of 30 \\
fev-bench-mini, win rate (points) & yes & 15.3 & 2.18 & 7.0 & 1 / 1 / 1 & -- & -- \\
Chronos Benchmark II, mean rank & yes & 4.24 & 0.22 & 19 & 0 / 0 / 1 & 8 & 3 of 11 \\
Chronos Benchmark II, geometric mean & no & 9.29 & 0.59 & 16 & 0 / 0 / 0 & 0 & 0 of 11 \\
TIME, geometric mean (log, displayed) & no & 0.0238 & 0.0021 & 11 & 0 / 1 / 1 & 0 & 0 of 44 \\
TIME, MASE rank column & yes & 4.01 & 0.378 & 11 & 1 / 1 / 1 & 13 & 12 of 44 \\
\midrule
BOOM, each series a source & yes & 0.196 & 0.312 & 0.63 & 9 / 9 / 9 & -- & -- \\
BOOM, 36 groups by domain and type & yes & 0.420 & 0.312 & 1.35 & 6 / 6 / 7 & & \\
BOOM, 23 groups by domain & yes & 0.635 & 0.312 & 2.04 & 5 / 6 / 7 & & \\
\bottomrule
\end{tabular}

\vspace{2pt}
\parbox{\textwidth}{\scriptsize\centering $^{a}$\,18 with tied ranks averaged. \ A dash:
history not replayed.}
\end{table}

Table~\ref{tab:allboards} gives every board under each of its rules. On Chronos
Benchmark~II win rate and mean rank give the same order. fev-bench-mini, with 20
of the 100 tasks, is twice as coarse as the full board (a resolution of 15.3 against 7.5
points) at the same gaps. BOOM resolves its neighbours when each of its 2{,}554 series
counts as a source. Grouped into 23 domains, five neighbouring pairs separate under CR2, and
one domain holds 44\% of the rows.

\paragraph{fev-bench's history.} fev-bench's repository holds 32 commits of results, and 30
updates could be replayed. Under win rate, with the board's own function, two updates
reversed a top-ten pair that nobody reran. In the update of 2026-10-13 to 10-14 the arrival
of timeh2o-v1-tiny reversed ChakraTS and TimesFM-3 at places 2 and 3 (win rates 85.39 and
85.26 before, 84.66 and 84.91 after). In the next update the arrival of TimeRSI reversed
TiRex-2 and Toto-2.0-2.5B at places 8 and 9 (73.22 and 73.06 before, 71.58 and 71.67 after).
On the current 40-entry board, removing one entry from outside the top ten reorders the top
ten in 2 of 30 cases. Adding a near-copy of an entry does so in 8 of 40.

\paragraph{TIME's history.} We scored every state of TIME's repository up to 2026-10-31 with
the files and the baseline as they were, on every entry present. Over the 44 updates the
MASE rank column reversed 13 untouched top-ten pairs in 12 updates, and the CRPS rank column
6 pairs in 5 updates. The two geometric means reversed none. Three of these events involve
two entries that were identical on 96 of 98 units for three minutes. With the opposite tie
order the counts move by at most one.

\paragraph{Chronos Benchmark~II with flagged results.} On this board 74 results are flagged
as trained on the test dataset. With them replaced by Chronos-Bolt's results, mean rank has
a ratio of 12, separates 1 of 9 neighbours and reverses 4 pairs in 2 of 11 updates. The
geometric mean then has a ratio of 23 and reverses none.

\paragraph{GIFT-Eval's sub-boards.} GIFT-Eval's app shows tabs by horizon, frequency, domain
and variate type. Ten of these groups hold at least 6 sources: short, medium and long
horizon, univariate and multivariate, hourly, minutely, daily and weekly, and the Econ/Fin
domain. The tabs have no default sort, so we order each group by mean rank on MASE within
the group. No sub-board separates a neighbouring pair under any procedure. The ratio of
resolution to median gap runs from 6.9 (short horizon, 55 configurations) to 43 (Econ/Fin, 6
configurations), against 7.5 for the full board. The leader changes on three of the ten
groups.

\paragraph{Newer models.} Timestral-209M \citep{timestral} is on three boards as read in
November 2026: 11th on GIFT-Eval, 3rd on TIME and 4th on fev-bench. No board separates it
from either neighbour, and its smallest $p$-value against a neighbour is 0.11.

\paragraph{New datasets and shared sources.} The figure of 48 times as many datasets assumes
that new datasets vary as the 28 do. Fifteen entries are scored on both GIFT-Eval and
fev-bench with the same checkpoint, which gives 105 pairs. On the 25 fev-bench families that
share no data source with GIFT-Eval, the spread of source means of the log error difference
is 1.13 times that on GIFT-Eval's 28 datasets (median over pairs, 95\% interval 0.85 to
1.72). The per-source noise scale that the calculation uses is 1.25 times as large (1.11 to
1.43). A ratio of 1.25 means about 1.6 times as many datasets again, so 48 is a lower
estimate.

The boards share some sources. Nine of fev-bench's 39 families share a data source with
GIFT-Eval by name, and 5 more may share one. With all 14 removed, 0 of 9 neighbours separate
and the resolution is 10.4 points against a median gap of 1.4. No TIME dataset shares a
source by name, and 8 cover the same topic from a different provider. With those removed,
0 of 9 separate under CR2 and the ratio is 16. Seven of Chronos Benchmark~II's 27 tasks
share a source with GIFT-Eval. The boards share entries more than data: 15 of BOOM's 23
entries match GIFT-Eval entries by name, and at least 15 of fev-bench's entries are on
GIFT-Eval.

\section{The history-length experiment}
\label{app:history}

\paragraph{Configurations.} A configuration is eligible when the median series of its data
holds at least 1{,}024 steps. A series shorter than 512 steps shows the model the same
history at both lengths. Of GIFT-Eval's 97 configurations 70 are eligible. They come from 18
base datasets, which we group into 14 sources: the three bizitobs datasets form one source,
the two ETT datasets another and the two bitbrains datasets a third. Twelve of the 70, from 9
sources, form the discovery set. A rule fixed before any model ran chose them: 5 short, 3
medium and 4 long horizons in proportion to the eligible counts, all from distinct base
datasets. The other 58 form the confirmation set.

\paragraph{Models.} The documented history length is the value in GIFT-Eval's reference
notebook for each model: 2{,}048 steps for Chronos-Bolt and 4{,}000 for Moirai-2.0. For
TimesFM-2.5 the notebook passes the full history up to 15{,}360 steps, and Toto receives the
smaller of 4{,}096 and the history available. The other setting gives every model the last
512 steps. Moirai-2.0's checkpoint reports a maximum length of 512, and its notebook passes
4{,}000. On the discovery set the two settings gave different forecasts, with a median
difference of 3.7\%. Chronos-Bolt, Moirai-2.0 and TimesFM-2.5 are deterministic: their 72
values on the discovery set repeat on a second machine to a relative difference of
$2.08\times10^{-5}$.

\paragraph{Effect and tests.} The effect of a model is the mean over configurations of the
log of its MASE at 512 steps over its MASE at the documented length, turned back into a
percentage. It equals the change in the geometric mean of scaled MASE. The first test asks
whether the effect differs from zero. It is an exact sign-flip test on the 14 source totals,
which lists all $2^{14}$ sign patterns. It is two-sided at 5\% with a Holm correction over
the four models, and the 95\% interval inverts it. The second test asks whether the effect
exceeds the spread of the displayed top ten on the same configurations. The spread is the
largest of the ten entries' geometric means of scaled MASE over the smallest, minus one. It is one-sided at
2.5\%. The estimate, both tests and their order were fixed before any run.

\begin{table}[!htb]
\caption{\textbf{Effect of cutting the history to 512 steps}, with the exact sign-flip test
on source totals. Each set of configurations has its own Holm correction. Toto's rows use
the reference notebook's setting and the first of three seeds.}
\label{tab:grid}
\centering
\footnotesize
\setlength{\tabcolsep}{6pt}
\begin{tabular}{@{}llrcrr@{}}
\toprule
Configurations & Model & Effect & 95\% interval & $p$ & $p$ after Holm \\
\midrule
All 70, 14 sources & Chronos-Bolt & +2.7\% & [$-$0.9, +6.2] & 0.1146 & 0.1146 \\
 & Moirai-2.0 & +6.2\% & [+1.6, +10.4] & 0.0051 & 0.0103 \\
 & TimesFM-2.5 & +15.3\% & [+4.1, +32.6] & 0.0002 & 0.0010 \\
 & Toto & +15.3\% & [+4.0, +27.1] & 0.0002 & 0.0010 \\
\midrule
Confirmation set, 58, & Chronos-Bolt & +3.7\% & [+0.3, +6.9] & 0.0354 & 0.0354 \\
13 sources & Moirai-2.0 & +6.1\% & [+1.3, +11.3] & 0.0085 & 0.0171 \\
 & TimesFM-2.5 & +14.0\% & [+4.7, +29.3] & 0.0002 & 0.0010 \\
 & Toto & +13.9\% & [+4.4, +24.3] & 0.0002 & 0.0010 \\
\midrule
Discovery set, 12, & Chronos-Bolt & $-$2.0\% & [$-$8.1, +4.6] & 0.9570 & 0.9570 \\
9 sources & Moirai-2.0 & +6.8\% & [+0.8, +12.1] & 0.0352 & 0.1055 \\
 & TimesFM-2.5 & +22.2\% & [$-$0.1, +55.9] & 0.0547 & 0.1094 \\
 & Toto & +22.0\% & [+1.8, +45.2] & 0.0156 & 0.0625 \\
\midrule
Discovery set, & TimesFM-3 & +31.8\% & [+5.3, +69.4] & 0.0156 & 0.0156 \\
newer models & Timestral-209M & +28.9\% & [+6.3, +54.1] & 0.0039 & 0.0078 \\
\bottomrule
\end{tabular}
\end{table}

\paragraph{Results.} On the 70 configurations the first test rejects after the Holm
correction for Moirai-2.0, TimesFM-2.5 and Toto, and it rejects for the same three on the
confirmation set (Table~\ref{tab:grid}). There it also rejects for Chronos-Bolt, at
$+$3.7\% with $p=0.035$. Chronos-Bolt does not separate on the 70, and the test claims a rise
too often for this model (below), so no effect is established for it. Of its 58 configurations, 43 come from a base
dataset that is also in the discovery set. On the discovery set alone no effect survives the
correction. Leaving one source out of the 70 moves the effect of TimesFM-2.5 between
$+$6.7\% and $+$17.7\%, that of Toto between $+$9.3\% and $+$16.7\% and that of Moirai-2.0
between $+$4.1\% and $+$6.9\%. The lowest values come from leaving out the bizitobs source.
The second test rejects for no model. The spread is 11.76\% on the 70, 12.71\% on the 58 and
9.34\% on the 12, and the intervals of TimesFM-2.5 and Toto start near 4\%.

\paragraph{Error rate of the test.} The sign-flip test assumes changes that are symmetric
when there is no effect, and these changes are skewed. We measured what the skew does. A
check centres each model's changes at zero, redraws them into the 14-source layout 20{,}000
times and counts the claims. For Moirai-2.0, TimesFM-2.5 and Toto, whose changes have their
long tail on the side of a rise, the test errs toward a fall. It claims a rise in at most
1\% of draws, against the 2.5\% that one side of a 5\% test allows, so it is conservative
for the rise we report. Its false claims go to the other side, where the two-sided rate is
12 to 19\%. TimesFM-2.5 and Toto also pass a test that uses directions only. For each of
them the error rises on 64 of the 70 configurations and in 13 of the 14 sources. A sign
test over the sources gives a one-sided $p$ of 0.0009. Chronos-Bolt's changes are skewed the
other way, and a false claim of a rise occurs for it in 4.75\% of draws, above the 2.5\%
allowed.

\paragraph{Newer models.} TimesFM-3 and Timestral-209M ran on the discovery set
with the same tests. Both reproduce the board's published rows at the
documented history length, to a relative difference of $1.5\times10^{-7}$. TimesFM-3's
error rises by 31.8\% at 512 steps and Timestral-209M's by 28.9\%, and both separate from
zero after a Holm correction over the two (Table~\ref{tab:grid}). The test against the
spread rejects for neither.

\paragraph{Sampling.} With no sample count Toto returns one point forecast. That setting
scores 0.8126 on the discovery set against 0.7699 from the board's published rows, and
single configurations are off by up to 28.55\%. The reference notebook's setting draws 256
samples and scores their median. It scores 0.7698, no configuration of the 70 is off by more
than 0.49\%, and three seeds give 0.7692 to 0.7709. That setting also turns on a key-value
cache and two numeric switches, so the comparison attributes the gap mainly to the sample
count and does not isolate it. Over the three seeds the effect on the discovery set runs
from $+$21.7\% to $+$22.1\%.

\begin{figure}[!htb]
\centering
\includegraphics[width=\linewidth,alt={Dot plot for four pretrained forecasting models, Chronos-Bolt-base, Moirai-2.0-R-small, Toto-Open-Base-1.0 and TimesFM-2.5-200M, of the share of the documented history length that each of 12 runs received. 13 of the 48 runs received less history than documented. 10, 9, 9 and 7 of 12 runs received the full length.}]{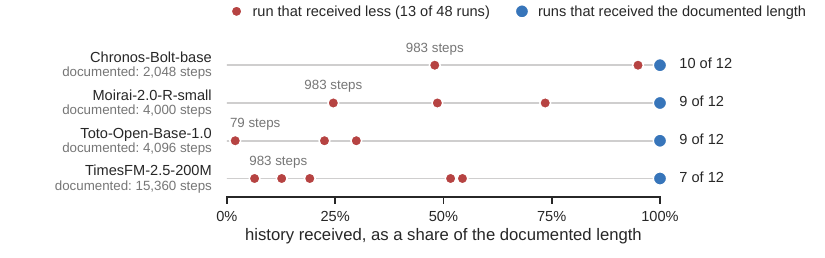}
\caption{\textbf{History received on the discovery set.} Each dot is one run at the
documented history length, placed by the share of that length the model received at the
median evaluation window. Red runs received less. Labels give the smallest history
received.}
\label{fig:received}
\end{figure}

\paragraph{History received.} At the median evaluation window, 13 of the 48 discovery-set
runs at a documented length received less history than documented (Figure~\ref{fig:received}).
The count is 14 by the minimum window and 12 by the maximum. On \texttt{us\_births}, models
with documented lengths of 2{,}048, 4{,}000 and 15{,}360 steps all received 983. Toto's
\texttt{m4\_daily} run received 79 steps at both lengths. The values are computed from the
series lengths and each model's rule. Datadog's benchmark chose 2{,}048
after preliminary experiments showed 512 degrading performance across models \citep{boom}.

\paragraph{Disclosure checklist.} The five fields of Table~\ref{tab:fields} are the ones that
differ between two honest evaluations of the same pretrained model, and none is recoverable
from the files boards publish today. The harness sets three of them, so a board can collect
them automatically.

\begin{table}[!htb]
\caption{\textbf{The five disclosure fields.} Each is a property of the evaluation run, not
of the model.}
\label{tab:fields}
\centering
\footnotesize
\begin{tabular}{@{}p{2.9cm}p{10.2cm}@{}}
\toprule
Field & What the entry reports \\
\midrule
History length & The number of past steps the model received per series, not the argument
it was passed. Report the median across evaluation windows and the minimum. \\
Aggregation rule & The statistic that turns per-unit scores into the displayed number,
including whether ranks are averaged before or after scaling. \\
Evaluation windows & The number of rolling windows per unit and their stride. \\
Normalisation & Whether inputs were scaled before the model saw them, and by which
statistic. \\
Sampling & The number of sample paths drawn and the summary that is scored. \\
\bottomrule
\end{tabular}
\end{table}

\end{document}